# Bayesian Multitask Learning with Latent Hierarchies

**Hal Daumé III**
School of Computing
University of Utah
Salt Lake City, UT 84112

## Abstract

We learn multiple hypotheses for related tasks under a latent hierarchical relationship between tasks. We exploit the intuition that for *domain adaptation*, we wish to share classifier structure, but for *multitask learning*, we wish to share covariance structure. Our hierarchical model is seen to subsume several previously proposed multitask learning models and performs well on three distinct real-world data sets.

## 1 INTRODUCTION

We consider two related, but distinct tasks: domain adaptation (DA) [4, 1, 7] and multitask learning (MTL) [5, 2]. Both involve learning related hypotheses on multiple data sets. In DA, we learn multiple classifiers for solving the *same problem* over data from *different distributions.* In MTL, we learn multiple classifiers for solving *different problems* over data from the *same distribution.*[1] Seen from a Bayesian perspective, a natural solution is a hierarchical model, with hypotheses as leaves [6, 16, 15]. However, when there are more than two hypotheses to be learned (i.e., more than two domains or more than two tasks), an immediate question is: are all hypotheses equally related? If not, what is their relationship? We address these issues by proposing two hierarchical models with *latent* hierarchies, one for DA and one for MTL (the models are nearly identical). We treat the hierarchy nonparametrically, employing Kingman's coalescent [12]. We derive an EM algorithm that makes use of recently developed efficient inference algorithms for the coalescent [14]. On several DA and MTL problems, we show the efficacy of our model.

Our models for DA and MTL share a common structure based on an unknown hierarchy. The key difference between the DA model and the MTL model is in what information is shared across the hierarchy. For simplicity, we consider the case of linear classifiers (logistic regression and linear regression). This can be extended to non-linear classifiers by moving to Gaussian processes [16]. In domain adaption, a useful model is to assume that there is a single classifier that "does well" on all domains [1, 15]. In the context of hierarchical Bayesian modeling, we interpret this as saying that the weight vector associated with the linear classifier is generated according to the hierarchical structure. On the other hand, in MTL, one does *not* expect the same weight vector to do well for all problems. Instead, a common assumption is that features co-vary in similar ways between tasks [13, 16]. In a hierarchical Bayesian model, we interpret this as saying that the covariance structure associated with the linear classifiers is generated according to the hierarchical structure. In brief: for DA, we share weights; for MTL, we share covariance.

## 2 BACKGROUND

### 2.1 RELATED WORK

Yu et al. [16] have presented a linear multitask model for domain adaptation. In the linear multitask model, a shared mean and covariance is generated by a Normal-Inverse-Wishart prior, and then the weight vector for each task is generated by a Gaussian conditioned on this shared mean and variance. The key idea in the linear multitask model [16] is to model feature covariance; this is also the intuition behind the informative priors model [13], carried out in a more Bayesian framework. (The linear multitask model is almost identical to the *conjoint analysis* model [6]).

[1] We note that this distinction is not always maintained in the literature where, often, DA is solved but it is called MTL. We believe this is valid (DA is a special case of MTL), but for the purposes of this paper, it is important to draw the distinction.

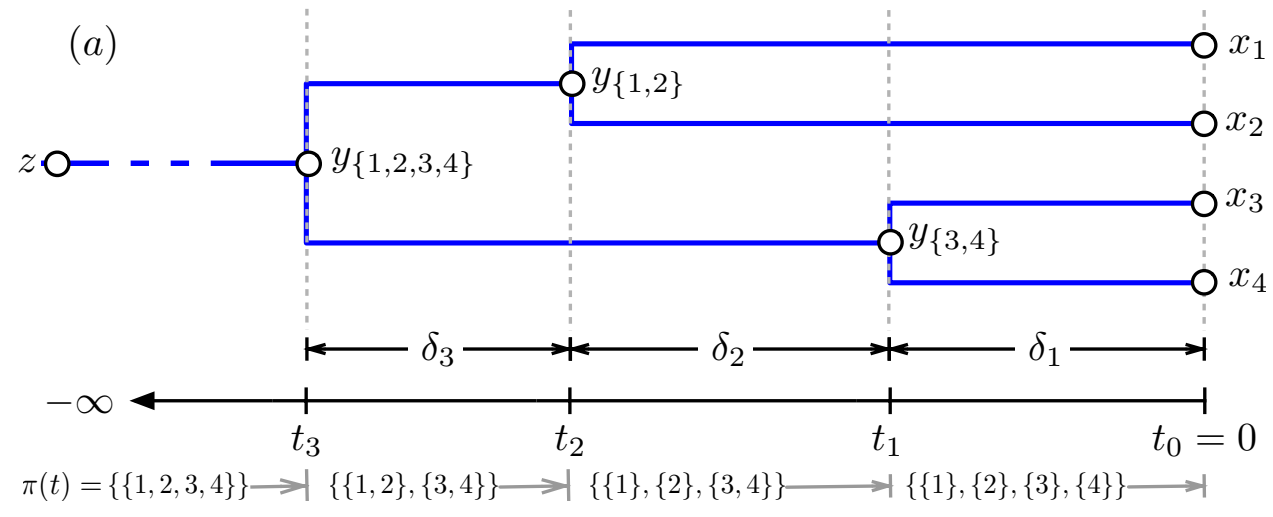

Figure 1: Variables describing the $N$-coalescent.

Xue et al. [15] present a Dirichlet process mixture model formulation, where domains are clustered into groups and share a single classifier across groups. This helps to prevent "negative transfer" (the effect of "unrelated" tasks negatively affecting performance on other tasks). Xue et al.'s model is effectively a *task-clustering* model, in which some tasks share common structure (those in the same cluster), but are otherwise independent from other tasks (those in other clusters). This work was later improved on by Dunson, Xue and Carin [9] in the formulation of the matrix stick breaking process: a more flexible approach to Bayesian multitask learning that allows for more sharing.

This is also a large body of work on non-Bayesian approaches to multitask learning and domain adaptation. Bickel et al. [2] offer an extension of the logistic regression model that simultaneously learns a good classifier and a classifier to provide instance weights for out of sample data. This approach is only applicable when no labeled "target" data is available, but much unlabeled target data is. Blitzer, McDonald and Pereira [4] present another approach to this "unsupervised" setting of domain adaptation that makes use of prior knowledge of features that are expected to behave similarly across domains. Both of these approaches are developed only in the two-domain setting. Dredze and Crammer [8] describe an online approach for dealing with the many-domains problem, sharing information across domains via confidence-weighted classifiers.

## 2.2 KINGMAN'S COALESCENT

Our model for DA and MTL makes use of a latent hierarchical structure. Being Bayesian, we wish to attach a prior distribution to this hierarchy. A convenient choice of prior is Kingman's coalescent [12]. Our description and notation is borrowed directly from [14]. Kingman's coalescent originated in the study of population genetics for a set of haploid organisms (organisms which have only a single parent). The coalescent is a nonparametric model over a countable set of organisms. It is most easily understood in terms of its finite dimensional marginal distributions over $N$ individuals, in which case it is called an $N$-coalescent. We then take the limit $N \to \infty$. In our case, the $N$ individuals will correspond to $N$ classifiers (tasks).

The $N$-coalescent considers a population of $N$ organisms at time $t = 0$ (see Figure 1 for an example with $N = 4$). We follow the ancestry of these individuals backward in time, where each organism has exactly one parent at time $t < 0$. The $N$-coalescent is a continuous-time, partition-valued Markov process which starts with $N$ singleton clusters at time $t = 0$ and evolves *backward*, coalescing lineages until there is only one left. We denote by $t_i$ the *time* at which the $i$th coalescent event occurs (note $t_i \leq 0$), and $\delta_i = t_i - t_{i-1}$ the time between events (note $\delta_i > 0$). Under the $N$-coalescent, each pair of lineages merges independently with exponential rate 1; so $\delta_i \sim \mathcal{E}xp\left(\binom{N-i+1}{2}\right)$. With probability one, a random draw from the $N$-coalescent is a binary tree with a single root at $t = -\infty$ and $N$ individuals at time $t = 0$. We denote by $\pi$ the tree structure and by $\boldsymbol{\delta}$ the collection of $\{\delta_i\}$. Leaves are denote by $x_n$ and internal nodes by $y_i$, where $i$ indexes a coalescent event (see Figure 1). The marginal distribution over tree topologies is uniform and independent of $\boldsymbol{t}, \boldsymbol{\delta}$; and the model is infinitely exchangeable. We consider the limit as $N \to \infty$, called *the coalescent.*

Once the tree structure is obtained, one can define an additional Markov process to evolve over the tree. One common, and easy to understand, choice is a Brownian diffusion process. In Brownian diffusion in $D$ dimensions, we assume an underlying diffusion covariance of $\boldsymbol{\Lambda} \in \mathbb{R}^{D \times D}$ positive semi-definite. The root is a $D$-dimensional vector drawn $\boldsymbol{z}$. Each $\boldsymbol{y}_i \in \mathbb{R}^D$ is drawn $\boldsymbol{y}_i \sim \mathcal{N}or(\boldsymbol{y}_{p(i)}, \delta_i \boldsymbol{\Lambda})$, where $p(i)$ is the parent of $i$ in the tree. $\boldsymbol{x}_i$s are drawn conditioned on their parent.

The coalescent is a very popular model in population genetics (it corresponds to a limiting case of the Wright-Fisher model), but has been plagued with the lack of efficient inference algorithm. (Most inference occurs by Metropolis-Hastings sampling over tree structures.) Recently, Teh et al. [14] proposed a collection of efficient bottom-up agglomerative inference algorithms for the coalescent. The one we make use is called Greedy-Rate1 and proceeds in a greedy manner, merging nodes that want to coalesce most quickly. In the case of Greedy-Rate1, the exponential rate is fixed as 1. Belief propagation is used to marginalize out internal nodes $y_i$. If we associate with each node in the tree a *mean* $y$ and *variance* $v$ message, we can compute messages as Eq (1), where $i$ is the current node and $li$ and $ri$ are its children.

$$\boldsymbol{v}_i = \left[(\boldsymbol{v}_{li} + (t_{li} - t_i)\boldsymbol{\Lambda})^{-1} + (\boldsymbol{v}_{ri} + (t_{ri} - t_i)\boldsymbol{\Lambda})^{-1}\right]^{-1} \tag{1}$$

$$\boldsymbol{y}_i = \left[\boldsymbol{y}_{li}(\boldsymbol{v}_{li} + (t_{li} - t_i)\boldsymbol{\Lambda})^{-1} + \boldsymbol{y}_{ri}(\boldsymbol{v}_{ri} + (t_{ri} - t_i)\boldsymbol{\Lambda})^{-1}\right]^{-1} \boldsymbol{v}_i$$

Importantly, this model is applicable when the $\boldsymbol{x}_i$s are not *known* entirely, but are represented by Gaussians. This can be done efficient since, given a hierarchical structure, inference is simply message passing in a Gaussian random field. (We will need this property in order to perform expectation-maximization.)

# 3 LATENT HIERARCHY MODELS

In this section, we present a model for domain adaptation (DA) and a model for multitask learning (MTL), plus some minor variants. (The variants are evaluated in Section 4.) As mentioned previously, the *structure* of the two models is the same: they differ in what information is shared.

To fix notation, suppose that we wish to learn $K$ different hypotheses ($K$ domains in DA or $K$ tasks in MTL). We suppose that we have training data for each hypothesis, with $N_k$ labeled examples examples for hypothesis $k$. (Notational confusion warning: in reference to the coalescent, the $K$ hypotheses will be the leaves of the coalescent tree, so this is more akin to a $K$-coalescent.) The inputs are drawn from $\mathbb{R}^D$ and outputs from $\mathcal{Y}$, where $\mathcal{Y} = \mathbb{R}$ for regression tasks or $\mathcal{Y} = \{-1, +1\}$ for classification tasks. We assume a distribution $\mathcal{D}^{(k)}$ over $\mathbb{R}^D$ for each hypothesis (in MTL, we assume identical distributions $\mathcal{D}^{(k)} = \mathcal{D}$). Our data thus has the form $\{\{(\boldsymbol{x}_n^{(k)}, y_n^{(k)}) : n \in [N_k]\} : k \in [K]\}$, where $[I] = \{1, \dots, I\}$, $\boldsymbol{x}_n^{(k)}$ is the $n$th input for task $k$ and $y_n^{(k)}$ is the corresponding label. Each $\boldsymbol{x}_n^{(k)} \sim \mathcal{D}^{(k)}$ iid. We will be using linear or logistic regression, parameterized by hypothesis-specific weight vectors $\boldsymbol{w}^{(k)} \in \mathbb{R}^D$, where predictions are made on the basis of $\boldsymbol{w}^{(k)\top}\boldsymbol{x}_n^{(k)}$.

One important design choice in both our models is whether we explicitly model the input $\boldsymbol{x}$. In the cases where we do *not*, our model is a conditional model of the form $p(y \mid \boldsymbol{x})$. In the cases where we *do*, our model is a joint model that factorizes as $p(y \mid \boldsymbol{x})p(\boldsymbol{x})$. In this case, the same tree structure is used to model both the conditional likelihood of $y$ given $\boldsymbol{x}$ *and* the data itself. In effect, this gives more data on which to learn the tree structure, at the cost that it might not be directly related to the prediction problem. We refer to this choice in the future as "model the data."

## 3.1 DOMAIN ADAPTATION

We propose the following model for domain adaptation. The basic idea is to generate a tree structure according to a $K$-coalescent and then propagate weight vectors along this tree. The root of the tree corresponds to the "global" weight vector and the leaves correspond to the task-specific weight vectors. We assume the weight vectors evolve according to Brownian diffusion. Our generative story is:

1. Choose a global *mean* and *covariance* $(\boldsymbol{\mu}^{(0)}, \boldsymbol{\Lambda}) \sim \mathcal{N}or\mathcal{IW}(0, \sigma^2\mathbf{I}, D+1)$. [2]
2. Choose a tree structure $(\pi, \boldsymbol{\delta}) \sim$ *Coalescent* over $K$ leaves.
3. For each non-root node $i$ in $\pi$ (top-down):
   (a) Choose $\boldsymbol{\mu}^{(i)} \sim \mathcal{N}or(\boldsymbol{\mu}^{(p_\pi(i))}, \delta_i\boldsymbol{\Lambda})$, where $p_\pi(i)$ is the parent of $i$ in $\pi$.
4. For each domain $k \in [K]$:
   (a) Denote by $\boldsymbol{w}^{(k)} = \boldsymbol{\mu}^{(i)}$ where $i$ is the leaf in $\pi$ corresponding to domain $k$.
   (b) For each example $n \in [N_k]$:
      i. Choose input $\boldsymbol{x}_n^{(k)} \sim \mathcal{D}^{(k)}$.
      ii. Choose output $y_n^{(k)}$ by:
      **Regression:** $\mathcal{N}or(\boldsymbol{w}^{(k)\top}\boldsymbol{x}_n^{(k)}, \rho^2)$
      **Classification:** $\mathcal{B}in(1/(1+e^{-\boldsymbol{w}^{(k)\top}\boldsymbol{x}_n^{(k)}}))$

Here, $\rho^2$ and $\sigma^2$ are hyperparameters that we assume are known (we use held-out data to set them).

We consider the following variants of this model: Is $\boldsymbol{\Lambda}$ is assumed diagonal or full? Do we explicitly model the data? We call these:

**Diag** Diagonal $\boldsymbol{\Lambda}$, do not model the data.

**Diag+X** Diagonal $\boldsymbol{\Lambda}$, do model the data.

**Full** Full $\boldsymbol{\Lambda}$, do not model the data.

**Full+X** Full $\boldsymbol{\Lambda}$, do model the data.

In the case where the input data is modeled explicitly (i.e., Diag+X and Full+X), we assume a base parameter vector over $\mathcal{X}$ generated at the root (in step (1)), propogated down the tree (in step (3)) and used to generate the inputs $\boldsymbol{x}_n^{(k)}$ (in step (4.b.i)). In the case that the input is modeled, we *always* assume diagonal covariance on the input. For continuous data, we use a Gaussian mutation kernel, as in step 4.a. For discrete data, we use a multinomial equilibrium distribution $\boldsymbol{q}_d$ and transition rate matrix $Q_d = \Lambda_{d,d}(\boldsymbol{q}_d{}^\top\mathbf{1}_K - \mathbf{I})$ where $\mathbf{1}_K$ is a vector of $K$ ones, while the transition probability matrix for entry $d$ in a time interval of length $\delta$ is $e^{Q_d t} = e^{-\delta\Lambda_{d,d}}\mathbf{I} + (1 - e^{-\delta\Lambda_{d,d}})\boldsymbol{q}_d{}^\top\mathbf{1}_K$.

[2] We denote by $\mathcal{N}or\mathcal{IW}(\boldsymbol{\mu}, \boldsymbol{\Lambda} \mid \boldsymbol{m}, \boldsymbol{\Psi}, \nu)$ the Normal-Inverse-Wishart distribution with prior mean $\boldsymbol{m}$, prior covariance $\boldsymbol{\Psi}$ and $\nu$ degrees of freedom.

### 3.2 MULTITASK LEARNING

In the multitask learning case, we no longer wish to share the weight vectors, but rather wish to share their *covariance* structure. This model is slightly more difficult to specify because Brownian motion no longer makes sense over a covariance structure (for instance, it will not maintain positive semi-definiteness). Our solution to this problem is to *decompose* the covariance structure into correlations and standard deviations. We assume a constant, global *correlation* matrix and only allow the standard deviations to evolve over the tree. (The idea of decomposing the covariance comes from [11], section 19.2.) We model the *log* standard deviations using Brownian diffusion.

In particular, our model assumes that each node in the tree is associated with a diagonal log standard deviation matrix $\mathbf{S}^{(i)} \in \mathbb{R}^{D \times D}$. The weight vector for task $k$ is then drawn Gaussian with zero mean and covariance given by $\big(\exp \mathbf{S}^{(i)}\big)\mathbf{R}\big(\exp \mathbf{S}^{(i)}\big)$, where $\mathbf{R} \in \mathbb{R}^{D \times D}$ are the shared correlations (with diagonal elements equal to 1). Our prior on $\mathbf{R}$ is:

$$p(\mathbf{R}) \propto (\det \mathbf{R})^{\frac{1}{2}(d+1)(d-1)-1} \prod_{i=1}^{D} (\det R_{(ii)})^{-(d+1)/2} \tag{2}$$

Here, $R_{(ii)}$ is the $i$th principle submatrix of $\mathbf{R}$. This is the marginal distribution of $\mathbf{R}$ when $\mathbf{SRS}$ has an inverse-Wishart prior with identity prior covariance and $D+1$ degrees of freedom, which leads to uniform marginals for each pairwise correlation.

Given this setup, our multitask learning model has the following generative story:

→ 1. Choose $\mathbf{R}$ by Eq (2) and deviation covariance $\mathbf{\Lambda} \sim \mathcal{IW}(\sigma^2 \mathbf{I}, D+1)$.
2. Choose a tree structure $(\pi, \boldsymbol{\delta}) \sim$ *Coalescent* over $K$ leaves.
3. For each non-root node $i$ in $\pi$ (top-down):
   → (a) Choose $\mathbf{S}^{(i)} \sim \mathcal{N}or(\mathbf{S}^{(p_\pi(i))}, \delta_i \mathbf{\Lambda})$, where $p_\pi(i)$ is the parent of $i$ in $\pi$.
4. For each task $k \in [K]$:
   → (a) Choose $\boldsymbol{w}^{(k)}$ by ($i$ is the leaf associated with task $k$): $\mathcal{N}or\big(0, \big(\exp \mathbf{S}^{(i)}\big)\mathbf{R}\big(\exp \mathbf{S}^{(i)}\big)\big)$
   (b) For each example $n \in [N_k]$:
      → i. Choose input $\boldsymbol{x}_n^{(k)} \sim \mathcal{D}$.
      ii. Choose output $y_n^{(k)}$ by:
      **Regression:** $\mathcal{N}or(\boldsymbol{w}^{(k)\top} \boldsymbol{x}_n^{(k)}, \rho^2)$
      **Classification:** $\mathcal{B}in(1/(1+e^{-\boldsymbol{w}^{(k)\top} \boldsymbol{x}_n^{(k)}}))$

The steps that *differ* from the the domain adaptation model are marked with an arrow (→).

### 3.3 INFERENCE

For both the DA and MTL models, we perform inference using an expectation-maximization algorithm. The *latent variables* in both algorithms are the variables associated with the leaves of the trees (in DA: the weight vectors; in MTL: the log standard deviations). The *parameters* are everything else: the tree structure $\pi$ and times $\boldsymbol{\delta}$, the Brownian covariance $\mathbf{\Lambda}$ and all other prior parameters.

#### 3.3.1 Domain adaptation

We begin with the domain adaptation model. For simplicity, we consider the case where the input data is *not* modeled. In the E-step, we compute expectations over the leaves (classifiers). In the M-step, we optimize the tree structure and the other hyperparameters.

**E-step:** The E-step can be performed exactly in the case of regression (the expectations of the classifiers are simply Gaussian). In the case of classification, we approximate the expectations by Gaussians (via the Laplace approximation). In particular, for each domain $k$, we compute:

$$\boldsymbol{w}^{(k)} = \arg\max_{\boldsymbol{w}} p(\boldsymbol{w}) \prod_{n=1}^{N_k} p(y_n^{(k)} \mid \boldsymbol{x}_n^{(k)}, \boldsymbol{w}) \tag{3}$$

$$\mathbf{C}^{(k)} = \left(\mathbf{X}^{(k)\top} \mathbf{A}^{(k)} \mathbf{X}^{(k)}\right)^{-1} + (\delta \mathbf{\Lambda})^{-1} \tag{4}$$

In Eq (3), $p(\boldsymbol{w})$ is the prior on $\boldsymbol{w}$ given by its parent in the tree; the likelihood term is the data likelihood (logistic for classification, or Gaussian for regression). We solve the optimization problem by conjugate gradient. $\boldsymbol{w}^{(k)}$ is the mean of the Gaussian representing the expectation of the $k$th weight vector. The covariance of the estimate is $\boldsymbol{C}^{(k)}$, with $\mathbf{A}^{(k)}$ diagonal. For regression, $\mathbf{A}^{(k)} = \mathbf{I}$; for classification, $\mathbf{A}^{(k)}$ has entries $A_{nn}^{(k)} = s_n^{(k)}(1 - s_n^{(k)})$, where $s_n^{(k)} = 1/(1 + e^{-\boldsymbol{w}^{(k)\top} \boldsymbol{x}_n^{(k)}})$.

**M-step:** Here, we optimize $(\pi, \boldsymbol{\delta})$ by integrating out $\boldsymbol{\mu}$s associated with internal nodes (using belief propagation). This can be done efficiently using the Greedy-Rate1 algorithm [14]. Optimize $\mathbf{\Lambda}$ as the mode of an Inverse-Wishart with $D+K+1$ degrees of freedom and mean $\mathbf{\Sigma}$:

$$\mathbf{\Sigma} = \mathbf{I} + \sum_i \boldsymbol{D}_i^\top \left(\boldsymbol{v}^{(l_\pi(i))} + \boldsymbol{v}^{(r_\pi(i))} + t^{(i)} \mathbf{\Lambda}\right)^{-1} \boldsymbol{D}_i \tag{5}$$

$$\boldsymbol{D}_i = \boldsymbol{\mu}^{(l_\pi(i))} - \boldsymbol{\mu}^{(r_\pi(i))} \quad , \quad t^{(i)} = \delta^{(l_\pi(i))} + \delta^{(r_\pi(i))} \tag{6}$$

Here, $l_\pi(i)$ and $r_\pi(i)$ are the left and right children respectively of node $i$ in $\pi$. $\boldsymbol{v}^{(i)}$ is the variance of node $i$ (obtained by Eq (4) for leaves or via belief propagation for internal nodes). The sum in Eq (5) ranges over all non-leaf nodes in $\pi$.

We initialize EM by computing $\boldsymbol{w}^{(k)}$ for each task according to a maximum a posteriori estimate with zero mean and $\sigma^2\mathbf{I}$ variance. This initialization effectively assumes no shared structure.

#### 3.3.2 Multitask learning

Constructing an exact EM algorithm for the multitask learning model is significantly more complex. The complexity arises from the convolution of the Normal (over $\boldsymbol{w}$) with the log-Normal (over $\boldsymbol{S}$). This makes the computation of exact expectations (over $\boldsymbol{S}$) intractable. We therefore use the popular "hard EM" approximation, in which we estimate the expectation of the latent variables ($\boldsymbol{S}$) with a point mass centered at their mode. (Experiments in the domain adaptation model show that the hard EM approximation to $\boldsymbol{w}$ does not affect results.)

The only additional complication is that of optimizing $\mathbf{R}$ (the overall correlations) and each $\mathbf{S}^{(i)}$ (the per-node standard deviations). $\mathbf{R}$ can be handled exactly as $\boldsymbol{\Lambda}$ in the domain adaptation case: see Eq (5), but constrained to have ones along the diagonal. The case for $\mathbf{S}^{(i)}$ is slightly more involved. We first maximize $\boldsymbol{w}$ as before, and then also maximize $\boldsymbol{S}$. The log posterior and its derivative have the forms below, where $C$ is a constant independent of $\mathbf{S}$ and $\mathbf{W} = \operatorname{diag} \boldsymbol{w}$:

$$\log p(\mathbf{S}) = -\operatorname{tr}\mathbf{S} - \frac{1}{2}\operatorname{tr}\left[(\mathbf{S}-\mathbf{P})^\top \boldsymbol{\Lambda}^{-1}(\mathbf{S}-\mathbf{P})\right] - \frac{1}{2}\operatorname{tr}\left[\mathbf{W}(e^{-\mathbf{S}}\mathbf{R}^{-1}e^{-\mathbf{S}})\mathbf{W}\right] + C$$

$$\nabla_{\mathbf{S}} \log p(\mathbf{S}) = -\mathbf{I} - (\mathbf{S}-\mathbf{P})\boldsymbol{\Lambda}^{-1} + \mathbf{W}(e^{-\mathbf{S}}\mathbf{R}^{-1}e^{-\mathbf{S}})\mathbf{W}$$

Here $\mathbf{P}$ is the (diagonal) matrix at the *parent* of the current node in the hierarchy. We optimize $\mathbf{S}$ by gradient descent with step size $(0.1/\mathit{iter})$ until convergence of $\mathbf{S}$ to $10^{-6}$.

## 4 EXPERIMENTAL RESULTS

We conduct experiments on two domain adaptation problems (sentiment analysis [3] and landmine detection [15]), and one multitask learning problem (based on a construction of 20-newsgroups previoulsy used for MTL [13]). The relevant dataset statistics for these data sets are in Table 1. Note that for both sentiment and 20-newsgroups, we project the data down to 50 dimensions using PCA. In all cases, we run EM for 20 iterations and choose the iteration for which the likelihood of 10% held-out training data is maximized.

Table 2: Performance on all tasks by competing models.

| Model | Sentiment N=100 | Sentiment N=6400 | Land-mine | 20 NG |
|---|---|---|---|---|
| **Indp** | 62.1% | 75.8% | 52.7% | 69.3% |
| **Pool** | 67.3% | 74.5% | 47.1% | - |
| **FEDA** | 63.6% | 75.7% | 51.6% | 69.5% |
| **YaXue** | 67.8% | 72.3% | 55.3% | 72.5% |
| **Bickel** | 68.0% | 72.5% | 55.5% | 74.1% |
| **Coal:** | | | | |
| **Full** | 72.2% | 80.5% | 56.2% | 75.8% |
| **Diag** | 71.9% | 80.4% | 55.8% | 75.3% |
| **Full+X** | 70.1% | 75.9% | 55.0% | 74.7% |
| **Diag+X** | 70.1% | 75.8% | 55.1% | 74.6% |
| **Data** | 70.1% | 75.8% | 54.9% | 72.0% |

For all experiments, we compare against the following baselines and alternative approaches:

**pool:** pool all the data and learn a single model

**indp:** train separate models for each domain/task

**feda:** the "augment" approach of by Daumé III [7]

**yaxue:** the flexible matrix stick breaking process method of Dunson, Xue and Carin [9]

**bickel:** the discriminative method of Bickel et al. [2][3]

The results for all data sets and all methods are shown in Table 2. Here, we also compare all five settings of the Coalescent model (full covariance and diagonal covariance, with and without the data, and then the tree derived just by clustering the data). Here, we can see that the more complex Coalescent-based models tend to outperform the other approaches.

### 4.1 DOMAIN ADAPTATION: SENTIMENT ANALYSIS

Our first experiment is on sentiment analysis data gathered from Amazon [3]. The task is to predict whether a review is positive or negative based on the text of the review. There are eight domains in this task: apparel (a), books (b), DVD (d), electronics (e), kitchen (k), music (m), video (v) and other (o). If we cluster these tasks on the basis of the *data*, we obtain the tree shown in Figure 2.

In our first experiment, we treat every domain equally and vary the amount of data used to learn a model. In

[3]The original method works only for two domains. We extend it to multiple domains in two ways: first, we do a one-versus-rest approach; second, we do a one-versus-one approach. The results presented here are *oracle* in the sense that they optimistically choose the better approach for each data set and each domain.

Table 1: Data set statistics for two DA problems and one MTL problem. The number of training and test examples are *averages* across the $K$ tasks and are presented with percentage standard deviation.

| Model | Dataset | # Tasks | # Features | # Train | # Test |
|---|---|---|---|---|---|
| **DA** | Sentiment [3] | 8 | 5964 | 9151±43% | 2288±43% |
| | Landmine detection [15] | 29 | 9 | 409±17% | 102±17% |
| **MTL** | 20-newsgroups [13] | 10 | 925 | 1127±8% | 751±8% |

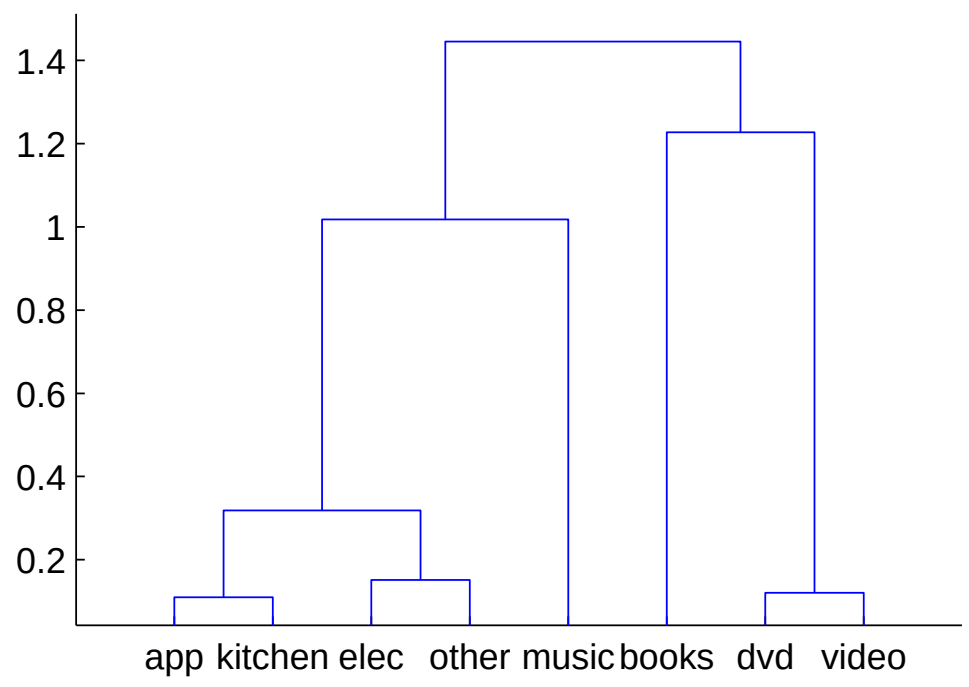

Figure 2: Coalescent tree obtained on sentiment data just using the data points.

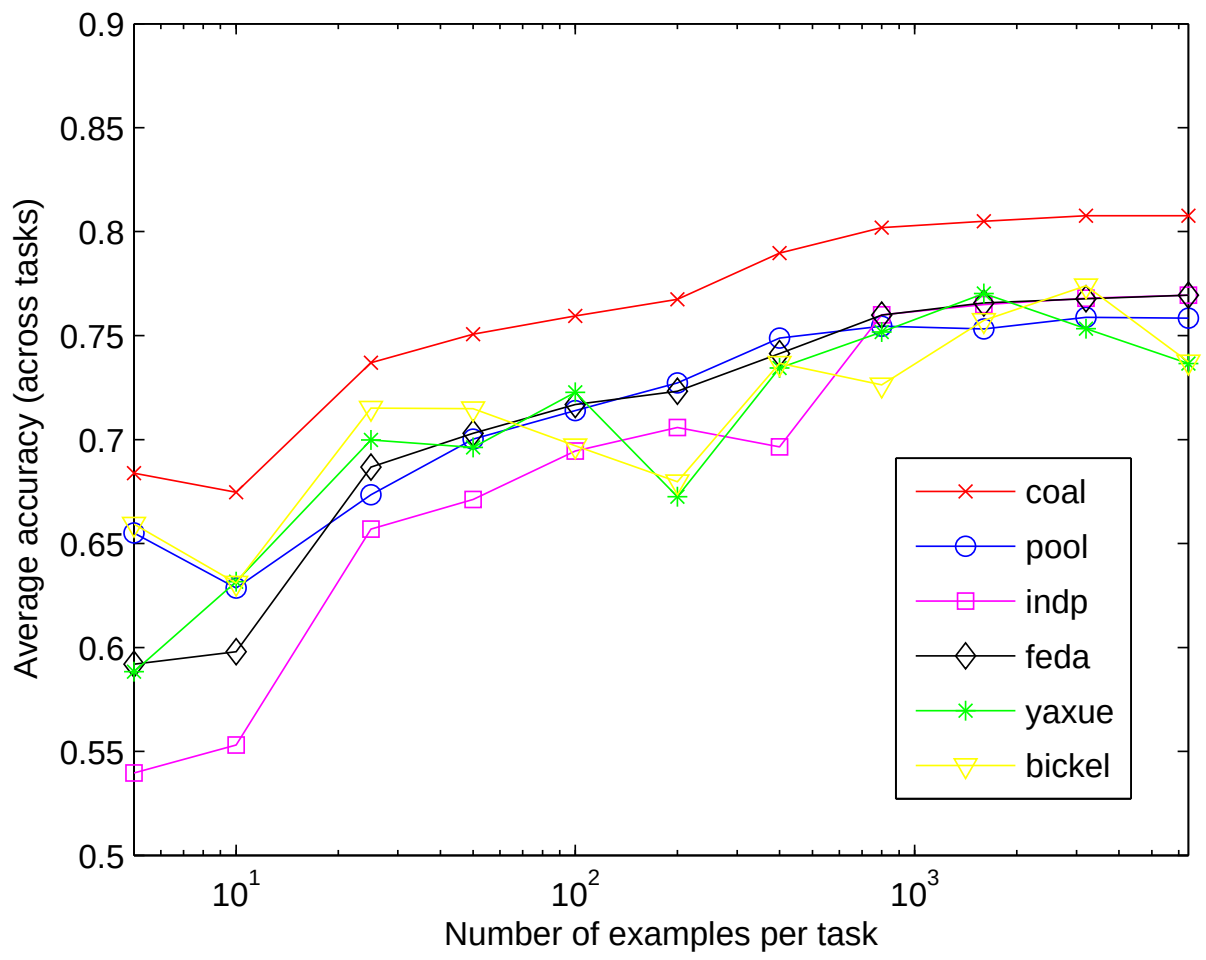

Figure 3: Accuracies on sentiment analysis data as number of data points per domain increases (coal = Full).

Figure 3, we show the results of the coalescent-based model (with full covariance but without data: Full), baselines, and comparison methods. As we can see, the coalescent-based approach dominates, even with very many data points (6400 per domain). In Table 2, we see that moving from full to diagonal covariance does not hurt significantly. Adding the data hurts performance significantly, and brings the performance down to the level of Data, the model that uses the data-based tree. In comparison to previously published results on this problem [3], our results are not quite as good. However, prior results depend on a large amount of prior knowledge in terms of "pivot features," which our model does not require, and also begin with a different feature representation.

In Figure 4, we show the trees after ten iterations of EM. We can see a difference between these trees and the tree built just on the data (cf., Figure 2). For instance, the data tree thinks that "music" is more like "appliances" than it is like "DVDs," something that does not happen in the EM tree.

In the next experiments, we select one task as the "target". We use 6400 examples from all the "source" tasks and vary the amount of labeled target data. We perform an evaluation on four targets, the same as those used previously [3]: books, DVD, electronics and kitchen. These results are shown in Figure 5. Here, we again see that the coalescent-based approach outperforms the baselines. However, for many of these per-target results, the feda baseline is the consistent-best alternative. One somewhat surprising result is that adding more and more target data does not appear to help significantly for this problem.

## 4.2 DOMAIN ADAPTATION: LANDMINE DETECTION

The second domain adaptation task we attempt is landmine detection [15]. To conserve space, we only present overall results and results for one subtask: the last one. To uncrowd the figure, we also limit the baseline models to a subset of approaches; recall that the full results are shown in Table 2. These are shown in Figure 6. Note that the performance measure here is AUC: there are very few positives in this data (around 5%). Here, we see that on the target-based evaluation, the coalescent-based approach dominates. For small amounts of data it performs equivalently to indp, but the gap increases for more data.

## 4.3 MULTITASK LEARNING: 20-NEWSGROUPS

Our final evaluation is on data drawn from 20-newsgroups. Here, we construct 10 binary classification problems, each of which is its own task. We use an identical setup to previous work [13]. As before, we present overall results and then results for two subtasks. The subtasks we choose are "Baseball versus Politics" and "IBM Hardware versus Forsale" – these were chosen as an example of good and bad transfer from previous studies [13]. Here, we have cut out the pool baseline because it does not make sense in a pure MTL setting. To uncrowd the figure, we also limit the baseline models to a subset of approaches; recall that the full results are shown in Table 2. The results are in Figure 7. Here, we see that the coalescent-based model overall outperforms the baselines, and further maintains an advantage for Baseball-versus-Politics, for which we expect a reasonable amount of transfer. One significant difference between these results and the DA results is that on the per-target results, in the DA case, our model continued to outperform. However, in the MTL case, with enough labeled target data, the independent classifiers quickly catch up. In comparison to prior results on this problem [13], our rate of improvement is roughly comparable.

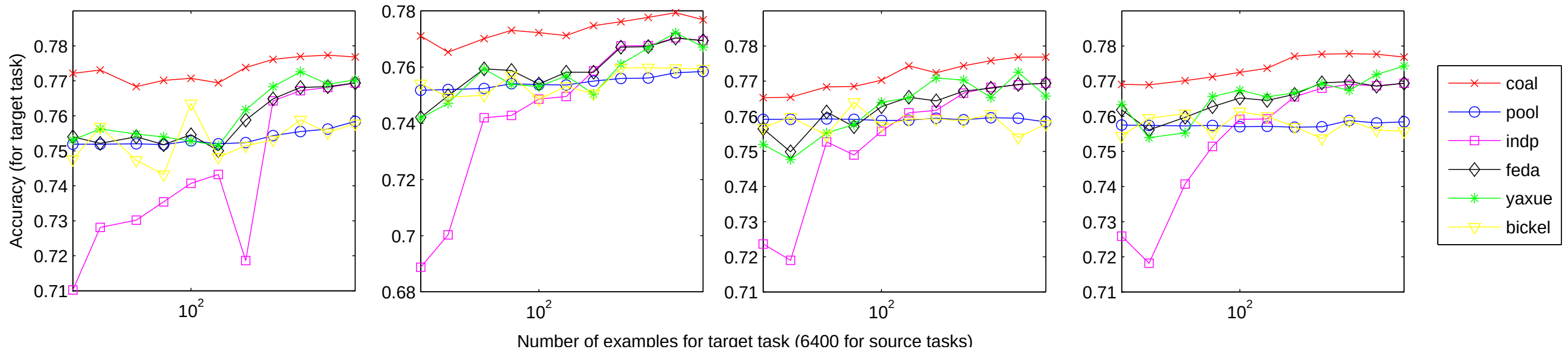

Figure 5: Per-target sentiment results.

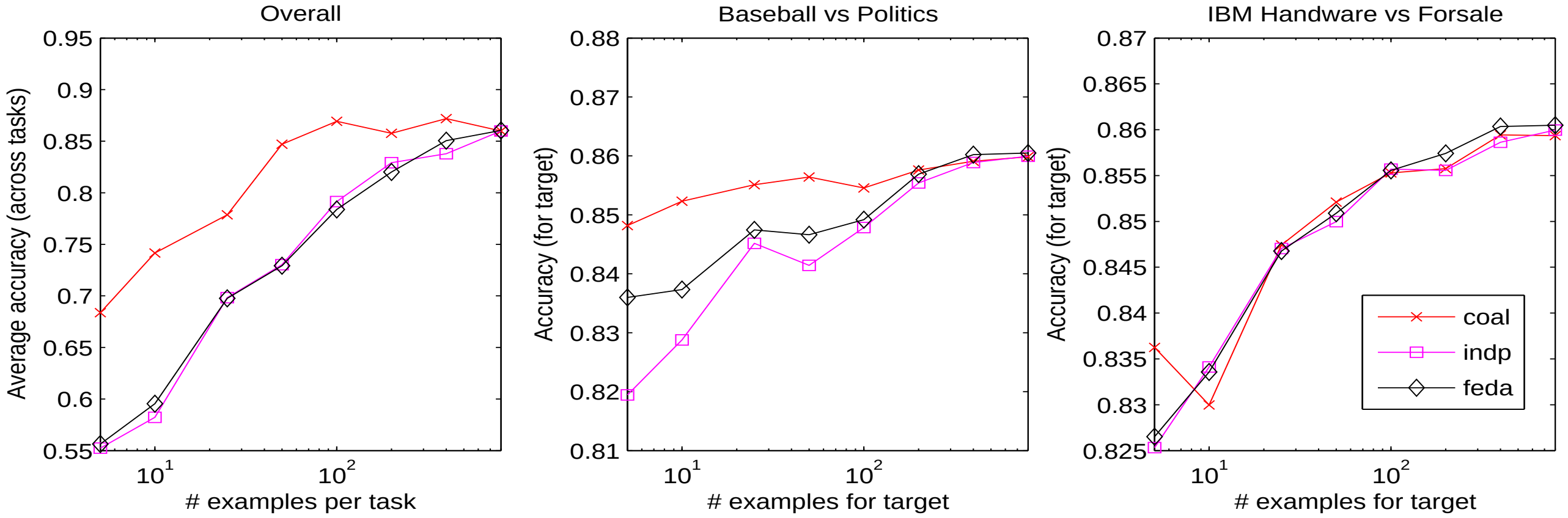

Figure 7: Results on 20-newsgroups multitask learning problem.

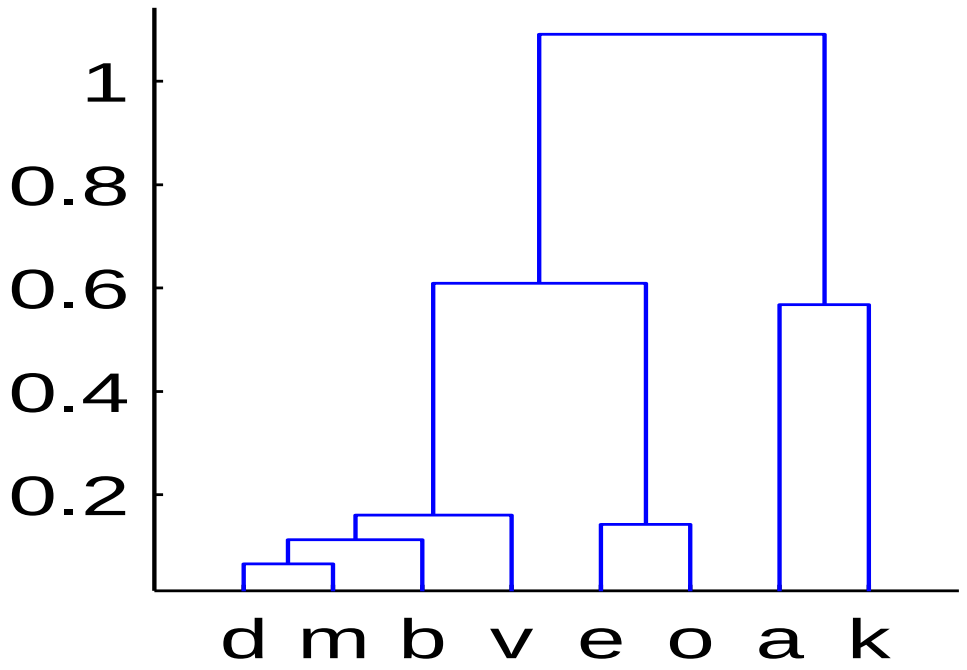

Figure 4: EM tree on the sentiment data.

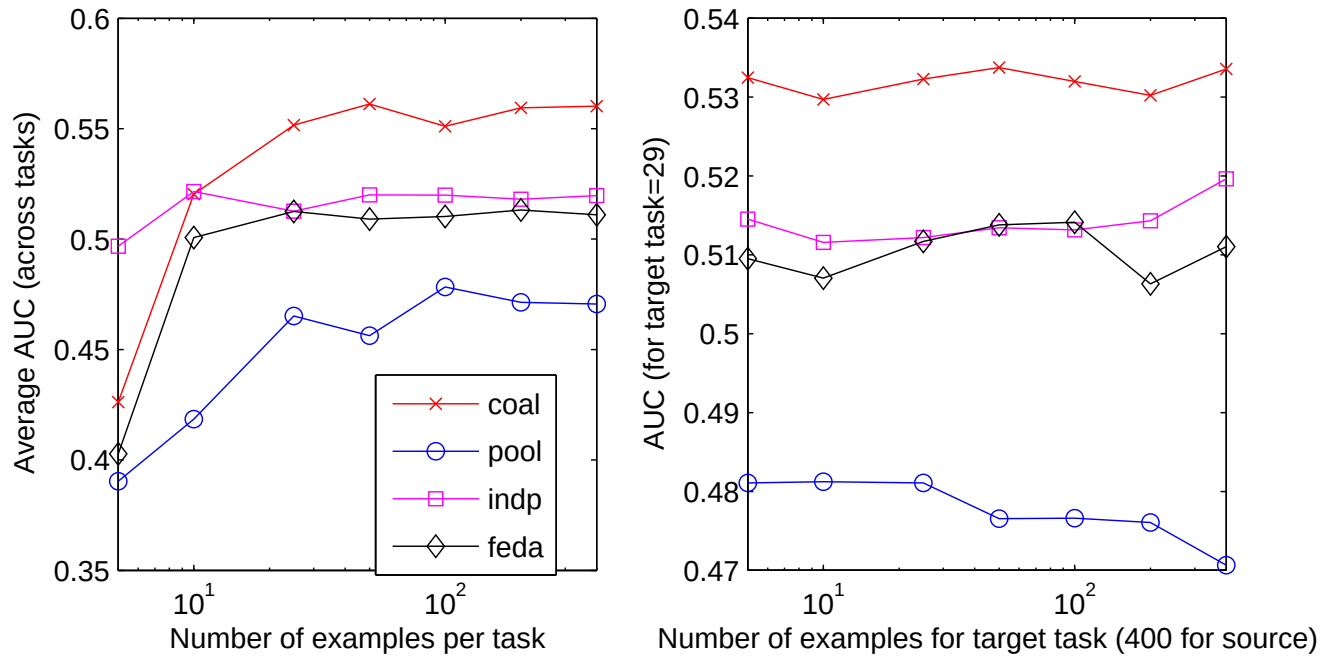

Figure 6: Landmine detection results.

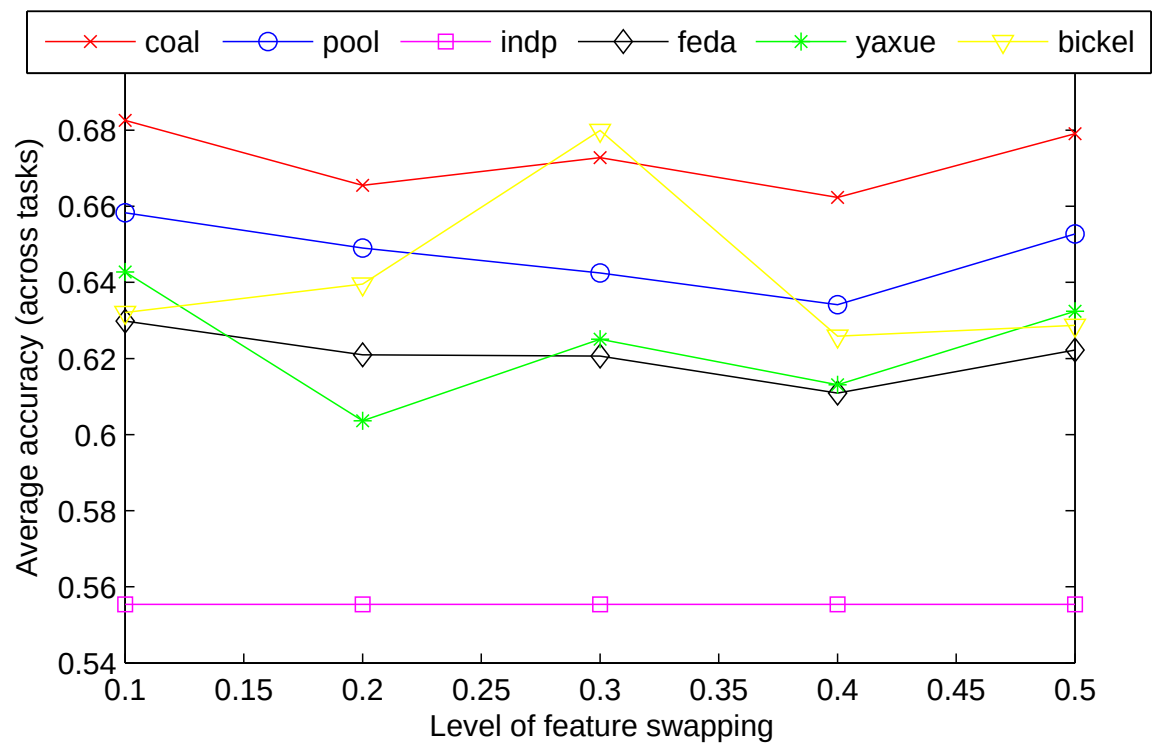

Figure 8: Adding bogus data to sentiment task.

### 4.4 RESULTS ON NOISY DOMAINS

One additional question that arises in work related to a large number of domains (or tasks) is whether the addition of unrelated domains can damage a model. In this section, we explore the effect of the addition of unrelated domains on our learning algorithms. We simulate this on the sentiment data by adding a task obtained by scrambling the features of one of the true tasks, where we vary the percentage scrambled.

The results are shown in Figure 8. Here, we see that there is a slighly trend toward degradation in performance for the original tasks as the amount of noise in the new task increases. This is true for all of the learning algorithms; unfortunately, this includes our own. One would hope that the model could learn to not share information with this irrelevant task, but apparently the prior toward short trees is too strong to overcome the noiser. Addressing this remains open.

## 5 DISCUSSION

We have presented two models: one for domain adaptation (DA) and one for multitask learning (MTL). Inference in our models is based on expectation maximization. We observe significant performance improvements on three very different data sets from our models. The only distinction between the models is what aspects are shared. We believe this is a reasonable way to divide up the DA/MTL landscape.

Two interesting special cases fall out of our model. First, if we set $\mathbf{\Lambda} = \mathbf{I}$ and construct a tree where every node branches directly from the root, our model is precisely the linear multitask model proposed by Yu et al. [16]. Second, we consider the fact that a special case of the coalescent can describe the same distribution as a *Dirichlet process* [10]. Through this view, we can see that Dirichlet-process based multitask model of Xue et al. [15] is achieved as a special case.

There are several ideas in the literature for both DA and MTL that are not reflected in our model. An easy example is the idea that it should be difficult to build a classifier for separating source from target data in a DA context [1]. Similar ideas have been exploited in discriminative models for domain adaptation [2]. However, these models are most successful when there is *no* labeled target data: a case we have not considered. It is an open question to address this in our framework.

**Acknowledgments.** We sincerely thank the many anonymous reviewers for helpful commentary. This was partially supported by NSF grant IIS-0712764.

## References

[1] S. Ben-David, J. Blitzer, K. Crammer, and F. Pereira. Analysis of representations for domain adaptation. *NIPS*, 2006.

[2] S. Bickel, M. Bruckner, and T. Scheffer. Discriminative learning for differing training and test distributions. *ICML*, 2007.

[3] J. Blitzer, M. Dredze, and F. Pereira. Biographies, bollywood, boom-boxes, and blenders: Domain adaptation for sentiment classification. *ACL*, 2007.

[4] J. Blitzer, R. McDonald, and F. Pereira. Domain adaptation with structural correspondence learning. *EMNLP*, 2006.

[5] R. Caruana. Multitask learning: A knowledge-based source of inductive bias. *Machine Learning*, 28:41–75, 1997.

[6] O. Chapelle and Z. Harchaoui. A machine learning approach to conjoint analysis. *NIPS*, 2005.

[7] H. Daumé III. Frustratingly easy domain adaptation. *ACL*, 2007.

[8] M. Dredze and K. Crammer. Online methods for multi-domain learning and adaptation. *EMNLP*, 2008.

[9] D. Dunson, Y. Xue, and L. Carin. The matrix stick-breaking process: flexible Bayes meta analysis. *JASA*, 103(481):317–327, 2008.

[10] T.S. Ferguson. A Bayesian analysis of some nonparametric problems. *Annals of Statistics*, 1(2):209–230, March 1973.

[11] A. Gelman, J. Carlin, H. Stern, and D. Rubin. *Bayesian Data Analysis*. Chapman & Hall/CRC, second edition, 2004.

[12] J. F. C. Kingman. On the genealogy of large populations. *Journal of Applied Probability*, 19:27–43, 1982. Essays in Statistical Science.

[13] R. Raina, A. Ng, and D. Koller. Constructing information priors using transfer learning. *ICML*, 2006.

[14] Y.W. Teh, H. Daumé III, and D. Roy. Bayesian agglomerative clustering with coalescents. *NIPS*, 2007.

[15] Y. Xue, X. Liao, L. Carin, and B. Krishnapuram. Multi-task learning for classification with Dirichlet process priors. *JMLR*, 2007.

[16] K. Yu, V. Tresp, and A. Schwaighofer. Learning Gaussian processes from multiple tasks. *ICML*, 2005.